\documentclass[11pt,a4paper]{article}
\usepackage[hyperref]{emnlp2020}
\usepackage{times}
\usepackage{latexsym}
\usepackage{graphicx}
\usepackage{amsmath}
\usepackage{amssymb}
\usepackage{bbm}
\usepackage{multirow}
\usepackage{tabularx}
\usepackage{ragged2e}
\usepackage{booktabs}

\usepackage{microtype}

\aclfinalcopy 
\def\aclpaperid{2268} 

\title{The Curse of Performance Instability in Analysis Datasets:\\Consequences, Source, and Suggestions}

\author{Xiang Zhou \ \ \ \ \ \ Yixin Nie \ \ \ \ \ \ Hao Tan \ \ \ \ \ \ Mohit Bansal \\
  Department of Computer Science \\
University of North Carolina at Chapel Hill \\
  \texttt{\{xzh, yixin1, haotan, mbansal\}@cs.unc.edu} \\
}

\date{}

\begin{document}
\maketitle

\begin{abstract}
We find that the performance of state-of-the-art models on Natural Language Inference (NLI) and Reading Comprehension (RC) analysis/stress sets can be highly unstable. 
This raises three questions: (1) How will the instability affect the reliability of the conclusions drawn based on these analysis sets? (2) Where does this instability come from? (3) How should we handle this instability and what are some potential solutions? 
For the first question, we conduct a thorough empirical study over analysis sets and find that in addition to the unstable final performance, the instability exists all along the training curve.
We also observe lower-than-expected correlations between the analysis validation set and standard validation set, questioning the effectiveness of the current model-selection routine.
Next, to answer the second question, we give both theoretical explanations and empirical evidence regarding the source of the instability, demonstrating that the instability mainly comes from high inter-example correlations within analysis sets.
Finally, for the third question, we discuss an initial attempt to mitigate the instability and suggest guidelines for future work such as reporting the decomposed variance for more interpretable results and fair comparison across models.\footnote{Our code is publicly available at: \url{https://github.com/owenzx/InstabilityAnalysis}}
\end{abstract}

\begin{figure}[t!]
    \centering
    \includegraphics[width=0.98\linewidth]{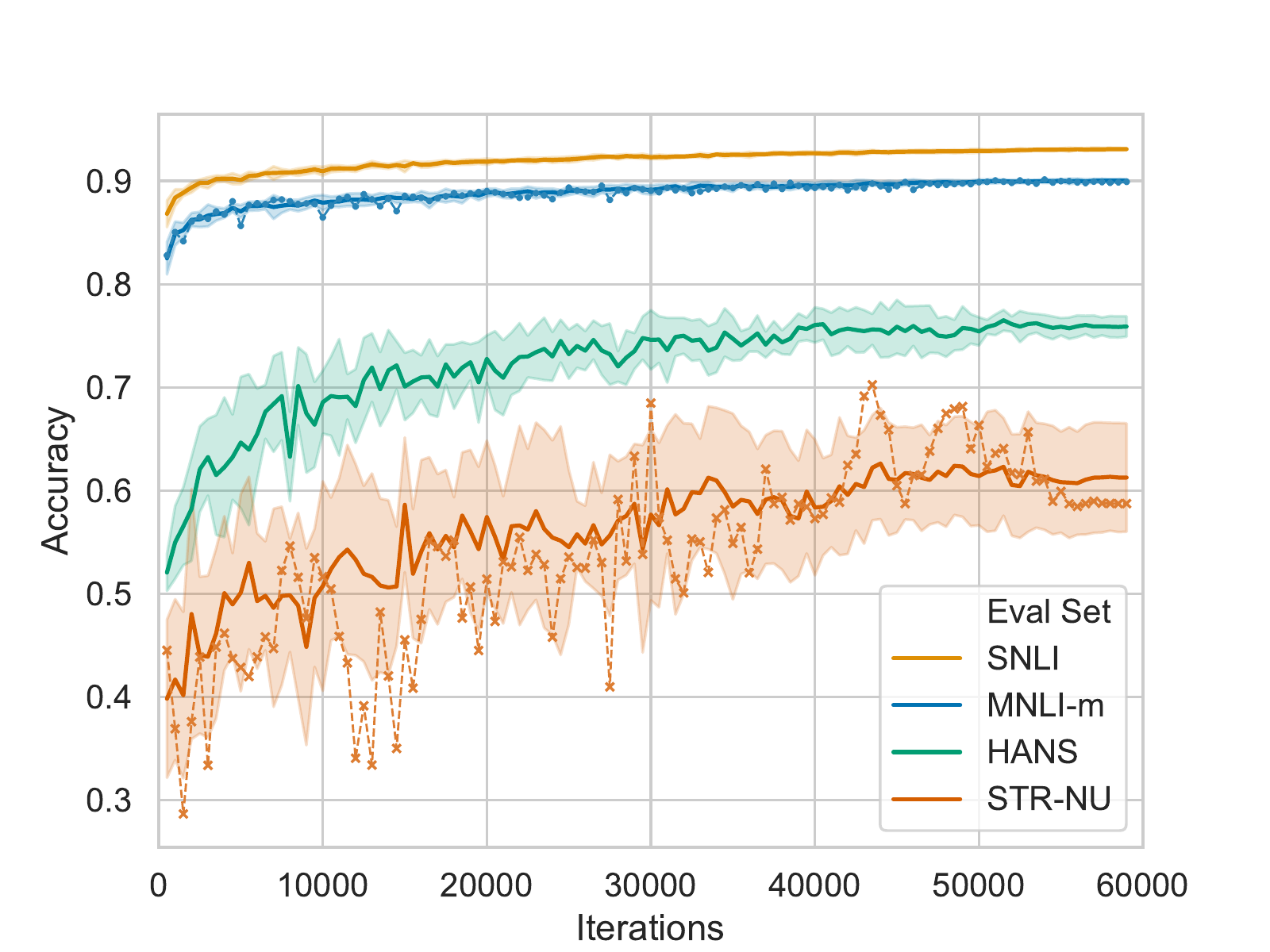}
    \caption{The trajectories of BERT performance on SNLI, MNLI-m, HANS~\cite{mccoy2019right}, and the Numerical subcategory of the Stress Test dataset~\cite{naik2018stress} (from the topmost line to the bottom, respectively). The solid lines represent the means of ten runs and the shadow area indicates a distance within a standard deviation from the means. The two dashed lines show the trajectories of one single run for MNLI-m and Numerical Stress Test using the same model.}
    \vspace{-5pt}
    \label{fig:bert_training_trajectory}
    \vspace{-5pt}
\end{figure}

\section{Introduction}
Neural network models have significantly pushed forward performances on natural language processing benchmarks with the development of large-scale language model pre-training~\cite{Peters_ELMO@2018, radford2018gpt_improving, devlin2019bert, radford2019gpt2_language, liu2019roberta}.  For example, on two semantically challenging tasks, Natural Language Inference (NLI) and Reading Comprehension (RC), the state-of-the-art results have reached or even surpassed the estimated human performance on certain benchmark datasets~\cite{wang2019glue, rajpurkar-etal-2016-squad, rajpurkar2018know}.
These astounding improvements, in turn, motivate a new trend of research to analyze what language understanding and reasoning skills are actually achieved, versus what is still missing within these current models.
Following this trend, numerous analysis approaches have been proposed to examine models' ability to capture different linguistic phenomena (e.g., named entities, syntax, lexical inference, etc.). Those studies are often conducted in 3 steps: (1) proposing assumptions about a certain ability of the model; (2) building analysis datasets by automatic generation or crowd-sourcing; (3) concluding models' ability using results on these analysis datasets. 

Past analysis studies have led to many key discoveries in NLP models, such as over-stability~\cite{jia2017advsquad}, surface pattern overfitting~\cite{gururangan2018annotation}, but recently \citet{mccoy2019berts} found that the results of different runs of BERT NLI models have large non-negligible variances on the HANS~\cite{mccoy2019right} analysis datasets, contrasting sharply with their stable results on standard validation set across multiple seeds. This finding raises concerns regarding the reliability of individual results reported on those datasets, the conclusions made upon these results, and lack of reproducibility~\cite{makel2012replications}.
Thus, to help consolidate further developments, we conduct a deep investigation on model instability, showing how unstable the results are, and how such instability compromises the feedback loop between model analysis and model development.

We start our investigation from a thorough empirical study of several representative models on both NLI and RC. Overall, we observe four worrisome observations in our experiments: (1) The final results of the same model with different random seeds on several analysis sets are of significantly \textbf{high variance}. The largest variance is more than 27 times of that for standard development set; (2) These large instabilities on certain datasets is \textbf{model-agnostic}. Certain datasets have unstable results across different models; (3) The instability not only occurs at the final performance but exists \textbf{all along training trajectory}, as shown in Fig.~\ref{fig:bert_training_trajectory}; (4) The results of the same model on analysis sets and on the standard development set have \textbf{low correlation}, making it hard to draw any constructive conclusion and questioning the effectiveness of the standard model-selection routine.

Next, in order to grasp a better understanding of this instability issue, we explore theoretical explanations behind this instability. Through our theoretical analysis and empirical demonstration, we show that inter-examples correlation within the dataset is the dominating factor causing this performance instability.
Specifically, the variance of model accuracy on the entire analysis set can be decomposed into two terms: (1) the sum of single-data variance (the variance caused by individual prediction randomness on each example), and (2) the sum of inter-data covariance (caused by the correlation between different predictions).
To understand the latter term better,  consider the following case: if there are many examples correlated with each other in the evaluation set, then the change of prediction on one example will influence predictions on all the correlated examples, causing high variances in final accuracy.
We estimate these two terms with multiple runs of experiments and show that inter-data covariance contributes significantly more than single-data variance to final accuracy variance, indicating its major role in the cause of instability.

Finally, in order for the continuous progress of the community to be built upon trustworthy and interpretable results, we provide initial suggestions on how to perceive the implication of this instability issue and how we should potentially handle it. For this, we encourage future research to: (1) when reporting means and variance over multiple runs, also report two decomposed variance terms (i.e., sum of single data variance and sum of inter-data covariance) for more interpretable results and fair comparison across models; (2) focus on designing models with better inductive and structural biases, and datasets with higher linguistic diversity.

Our contribution is 3-fold. First, we provide a thorough empirical study of the instability issue in models' performance on analysis datasets. Second, we demonstrate theoretically and empirically that the performance variance is attributed mostly to inter-example correlations. Finally, we provide suggestions on how to deal with instability, including reporting the decomposed variance for more interpretable evaluation and better comparison.

\section{Related Work}

\paragraph{NLI and RC Analysis.}
 Many analysis works have been conducted to study what the models are actually capturing alongside recent improvements on NLI and RC benchmark scores. In NLI, some analyses target word/phrase level lexical/semantic inference \cite{glockner2018breaking, shwartz2018paraphrase, carmona2018behavior}, some are more syntactic-related \cite{mccoy2019right, nie2019analyzing, geiger2019posing}, some also involved logical-related study \cite{minervini2018adversarially, wang2019glue} and some involved pragmatic inference~\cite{jeretic-etal-2020-natural}. \newcite{naik2018stress} proposed a suite of analysis sets covering different linguistic phenomena.  \newcite{ynie2020chaosnli} studied the factor of collective human opinion distributions on model performance which might also have connections with performance instability. In RC, adversarial style analysis is used to test the robustness of the models~\cite{jia2017advsquad}.
Most of the work follows the style of \citet{carmona2018behavior} to diagnose/analyze models' behavior on pre-designed analysis sets.
In this paper, we analyze NLI and RC models from a broader perspective by inspecting models' performance across different analysis sets, and their inter-dataset and intra-dataset relationships.

\paragraph{Dataset-Related Analysis.}
Another line of works study the meta-issues of the dataset. The most well-known one is the analysis of undesirable bias. In VQA datasets, unimodal biases were found, compromising their authority on multi-modality evaluation~\cite{jabri2016revisiting, goyal2017making}. In RC, \newcite{kaushik2018much} found that passage-only models can achieve decent accuracy.
In NLI, hypothesis bias was also found in SNLI and MultiNLI~\cite{tsuchiya2018performance, gururangan2018annotation}.
These findings revealed the spurious shortcuts in the dataset and their harmful effects on trained models.
To mitigate these problems, 
\newcite{liu2019inoculation} introduced a systematic task-agnostic method to analyze datasets. \newcite{rozen2019diversify} further explain how to improve challenging datasets and why diversity matters. \newcite{geva2019we} suggest that the training and test data should be from exclusive annotators to avoid annotator bias. Our work is complementary to those analyses.

\paragraph{Robustifying NLI and RC Models.}
Recently, a number of works have been proposed to directly improve the performance on the analysis datasets both for NLI through model ensemble~\cite{clark2019don, he2019unlearn}, novel training mechanisms~\cite{pang2019improving, yaghoobzadeh2019robust}, adversarial data augmentation~\cite{nie2019adversarial}, enhancing word representations~\cite{moosavi2019improving}, and for RC through different training objectives ~\cite{yeh2019qainfomax, Lewis2019GenerativeQA}. While improvements have been made on certain analysis datasets, the stability of the results is not examined. As explained in this paper, we highly recommend those result variances be scrutinized in future work for fidelity considerations.

\paragraph{Instability in Performance.}
Performance instability has already been recognized as an important issue in deep reinforcement learning~\cite{rlblogpost} and active learning~\cite{bloodgood2013analysis}. However, supervised learning is presumably stable especially with fixed datasets and labels. This assumption is challenged by some analyses recently. \newcite{mccoy2019berts} show high variances in NLI-models performance on the analysis dataset. \newcite{phang2018sentence} found high variances in fine-tuning pre-trained models in several NLP tasks on the GLUE Benchmark. \newcite{reimers2017reporting, reimers2018comparing} state that conclusions based on single run performance may not be reliable for machine learning approaches. \newcite{weber2018fine} found that the model's ability to generalize beyond the training distribution depends greatly on the random seed. \newcite{dodge2020fine} showed weight initialization and training data order both contribute to the randomness in BERT performance. \newcite{nie-etal-2020-simple} found that combining training data from different tasks in multi-tasking training setting also induces instability in the training trajectory. In our work, we present a comprehensive explanation and analysis of the instability of neural models on analysis datasets and give general guidance for future work.

\begin{figure*}[t!]
    \centering
    \includegraphics[width=0.98\linewidth]{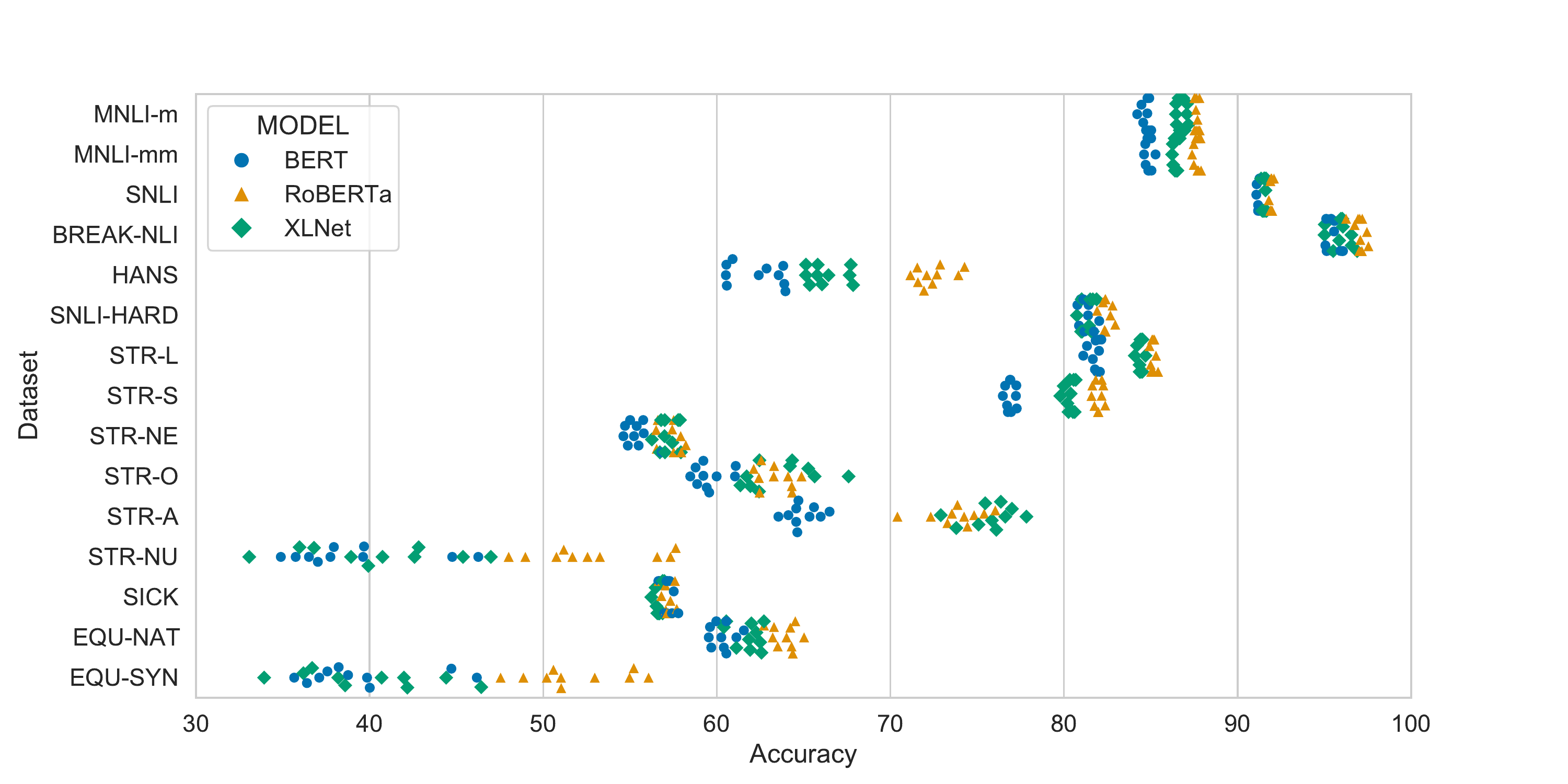}
    \vspace{-6pt}
    \caption{The results of BERT, RoBERTa, and XLNet on all datasets with 10 different random seeds. Large variance can be seen at certain analysis datasets (e.g. STR-NU, HANS, etc.) while results on standard validation sets are always stable.}
    \vspace{-9pt}
    \label{fig:bert_results}
\end{figure*}
\section{The Curse of Instability}

\subsection{Tasks and Datasets}
In this work, we target our experiments on NLI and RC for two reasons: 1) their straightforwardness for both automatic evaluation and human understanding, and 2) their wide acceptance of being benchmarks evaluating natural language understanding.

For NLI, we use {SNLI}~\cite{bowman2015large} and {MNLI}~\cite{williams2018broad} as the main standard datasets and use {HANS}~\cite{mccoy2019right}, {SNLI-hard}~\cite{gururangan2018annotation}, {BREAK-NLI}~\cite{glockner2018breaking}, {Stress Test}~\cite{naik2018stress}, {SICK}~\cite{marelli2014sick}, {EQUATE}~\cite{ravichander2019equate} as our auxiliary analysis sets. Note that the {Stress Test} contains 6 subsets (denoted as `STR-X') targeting different linguistic categories. We also splite the {EQUATE} dataset to two subsets (denoted as `EQU-NAT/SYN') based on whether the example are from natural real-world sources or are controlled synthetic tests.
For RC, we use {SQuAD1.1}~\cite{rajpurkar2016squad} as the main standard dataset and use {AdvSQuAD}~\cite{jia2017advsquad} as the analysis set. All the datasets we use in this paper are English. Detailed descriptions of the datasets are in Appendix. 

\subsection{Models}
Since BERT~\cite{devlin2019bert} achieves state-of-the-art results on several NLP tasks, the pretraining-then-finetuning framework has been widely used. To keep our analysis aligned with recent progress, we focused our experiments on this framework. Specifically, in our study, we used the two most typical choices: BERT~\cite{devlin2019bert} and XLNet~\cite{yang2019xlnet}.\footnote{For all the transformer models, we use the implementation in \url{https://github.com/huggingface/transformers}. BERT-B, BERT-L stands for BERT-base and BERT-large, respectively. The same naming rule applies to other transformer models.} Moreover, for NLI, we additionally use RoBERTa~\cite{liu2019roberta} and ESIM~\cite{chen2017enhanced} in our experiments. RoBERTa is almost the same as BERT except that it has been trained on 10 times more data during the pre-training phrase to be more robust. ESIM is the most representative pre-BERT model for sequence matching problem and we used an ELMo-enhanced-version~\cite{Peters_ELMO@2018}.\footnote{For ESIM, we use the implementation in AllenNLP~\cite{Gardner2017AllenNLP}.}
All the models and training details are in Appendix.

\subsection{What are the Concerns?}
\paragraph{Instability in Final Performance.} 
Models' final results often serve as a vital measurement for comparative study. Thus, we start with the question: ``How unstable are the final results?''
To measure the instability, we train every model $10$ times with different random seeds. Then, we evaluate the performances of all the final checkpoints on each NLI dataset and compute their standard deviations.
As shown in Fig.~\ref{fig:bert_results}, the results of different runs for BERT, RoBERTa, and XLNet are highly stable on MNLI-m, MNLI-mm, and SNLI, indicating that model performance on standard validation datasets regardless of domain consistency\footnote{Here SNLI and MNLI-m share the same domain as the training set while MNLI-mm is from different domains.} are fairly stable.
This stability also holds on some analysis sets, especially on SNLI-hard, which is a strict subset of the SNLI validation set. On the contrary, there are noticeable high variances on some analysis sets. The most significant ones are on STR-NU and HANS where points are sparsely scattered, with a 10-point gap between the highest and the lowest number for STR-NU and a 4-point gap for HANS.

\paragraph{Model-Agnostic Instability.}
Next, we check if the instability issue is model-agnostic. For a fair comparison, as the different sizes of the datasets will influence the magnitude of the instability, we normalize the standard deviation on different datasets by multiplying the square root of the size of the dataset\footnote{This normalization factor assumes that every prediction is independent of each other.} and focus on the relative scale compared to the results on the MNLI-m development set, i.e., $\frac{STD(dataset)}{STD(MNLI-m)}\sqrt{\frac{SIZE(dataset)}{SIZE(MNLI-m)}}$. The results for all the models are shown in Table~\ref{tab:all_instability} (the original means and standard deviations are in Appendix). 
From Table~\ref{tab:all_instability}, we can see that the instability phenomenon is consistent across all the models. Regardless of the model choice, some of the analysis datasets (e.g., HANS, STR-O, STR-N) are significantly more unstable (with standard deviation 27 times larger in the extreme case) than the standard evaluation datasets. 
Similarly, for RC, the normalized deviation of model F1 results on SQuAD almost doubled when evaluated on AddSent, as shown in Table~\ref{tab:squad_instability} (the original means and standard deviations are in Appendix).

\begin{table*}[t]
\resizebox{\textwidth}{!}{%
\begin{tabular}{lccccccccccccccc}
\toprule
\multirow{2}{*}{\textbf{Model}}   & \multicolumn{3}{c}{Standard Datasets}                                                  & \multicolumn{12}{c}{Analysis Sets}    \\ 
\cmidrule(lr){2-4}  \cmidrule(lr){5-16} 
         & \bf  MNLI-m & \bf MNLI-mm & \bf SNLI &  \bf BREAK-NLI & \bf HANS  & \bf SNLI-hard & \bf STR-L & \bf STR-S & \bf  STR-NE & \bf STR-O & \bf STR-A & \bf STR-NU &\bf SICK &\bf EQU-NAT &\bf EQU-SYN \\
\midrule
\textbf{ESIM}          & 1.00     &  0.57        &   0.73  &  \underline{3.84}     &  0.82          & 0.73        &   0.77    &  0.73     & 3.57      &  \bf 4.63     &   2.58    &  2.79  & 1.47 & 1.19 & 2.70   \\
\textbf{ESIM+ELMo}     &  1.00    &  2.00        &  1.50  &    \underline{11.5}   &    4.55        &   2.48      &  3.10     &  2.20     &  7.50     & \bf 15.5     &  6.38     &   8.36   & 2.28 & 2.36 & 8.45  
\\
\textbf{BERT-B}     &    1.00  &    0.83      & 0.48    &    1.43     & 10.95          & 0.95        &  1.39     &   1.04    &   2.70    &   3.70    &   1.46    &  \bf  13.65  & 1.48 & 1.03 & \underline{13.17}  \\
\textbf{RoBERTa-B}  &    1.00  &    1.46      & 0.64     &   2.82     & 15.42          & 1.47        &  1.27     &   2.17    &  5.45      &   8.45    &   5.55    & \bf   25.75  & 2.91 & 2.29 & \underline{22.68}  \\
\textbf{XLNet-B}    & 1.00  &    0.48      & 0.37 &   2.03     &  6.60         &  0.75       &    0.59   &    0.92   &    1.96    &    7.19   &    2.07   &  \underline{13.33}  & 0.82 & 1.15 & \bf{13.33} \\
\textbf{BERT-L}  & 1.00 & 1.13 & 0.56 &     2.86 &     18.47 & 1.37 & 1.31 &    2.63 & 9.19 & 10.13 & 2.39 & \bf 21.88 & 1.71 & 1.41 & \underline{20.36} \\
\textbf{RoBERTa-L} & 1.00 & 0.88 & 0.69 &  1.03 & 10.27 & 1.01 & 1.12 & 1.20 & 12.13 & 10.13 & 4.51 & \bf    27.38 & 1.71 & 1.21 & \underline{22.36} \\
\textbf{XLNet-L}   &    1.00 &    0.90 &    0.69 &     1.06 &    10.67 &     0.85 & 0.89 & 1.45 &  \underline{16.21} &    11.84 & 4.26 &  15.93 & 1.50 & 1.31 &\bf 19.93  \\
\bottomrule
\end{tabular}%
}
\caption{Relatively normalized deviations of the results on MNLI-m for all models. The highest deviations are in bold and the second highest deviations are underlined for each individual model.}
\vspace{-5pt}

\label{tab:all_instability}
\end{table*}

\begin{table}[t]
\resizebox{0.47\textwidth}{!}{%
\begin{tabular}{lccc}
\toprule
\multirow{2}{*}{\bf Model}   & \multicolumn{1}{c}{Standard Dataset}                                                  & \multicolumn{2}{c}{Analysis Sets}    \\ 
\cmidrule(lr){2-2}  \cmidrule(lr){3-4} 
         &   \bf SQuAD &  \bf AddSent & \bf AddOneSent \\
\midrule
\textbf{BERT-B}     &    1.00  & \bf 2.61 & 1.58      \\
\textbf{XLNet-B}    &    1.00  &  \bf 1.78  &    1.00    \\
\bottomrule
\end{tabular}%
}
\caption{Relatively normalized deviations of the results on SQuAD dev set for both BERT-B and XLNet-B.}
\vspace{-5pt}
\label{tab:squad_instability}
\end{table}

\paragraph{Fluctuation in Training Trajectory.}
\label{sec:instable_train} 
Intuitively, the inconsistency and instability in the final performance of different runs can be caused by the randomness in initialization and stochasticity in training dynamics. To see how much these factors can contribute to the inconsistency in the final performance, we keep track of the results on different evaluation sets along the training process and compare their training trajectories. We choose HANS and STR-NU as our example unstable analysis datasets because their variances in final performance are the largest, and we choose SNLI and MNLI-m for standard validation set comparison.
As shown in Fig.~\ref{fig:bert_training_trajectory}, the training curve on MNLI and SNLI (the top two lines) is highly stable, while there are significant fluctuations in the HANS and STR-NU trajectories (bottom two lines). 
Besides the mean and standard deviation over multiple runs, we also show the accuracy of one run as the bottom dashed line in  Fig.~\ref{fig:bert_training_trajectory}. We find that two adjacent checkpoints can have a dramatically large performance gap on STR-NU. Such fluctuation is very likely to be one of the reasons for the instability in the final performance and might give rise to untrustworthy conclusions drawn from the final results.

\begin{figure}[t]
    \centering
    \includegraphics[width=1.0\linewidth]{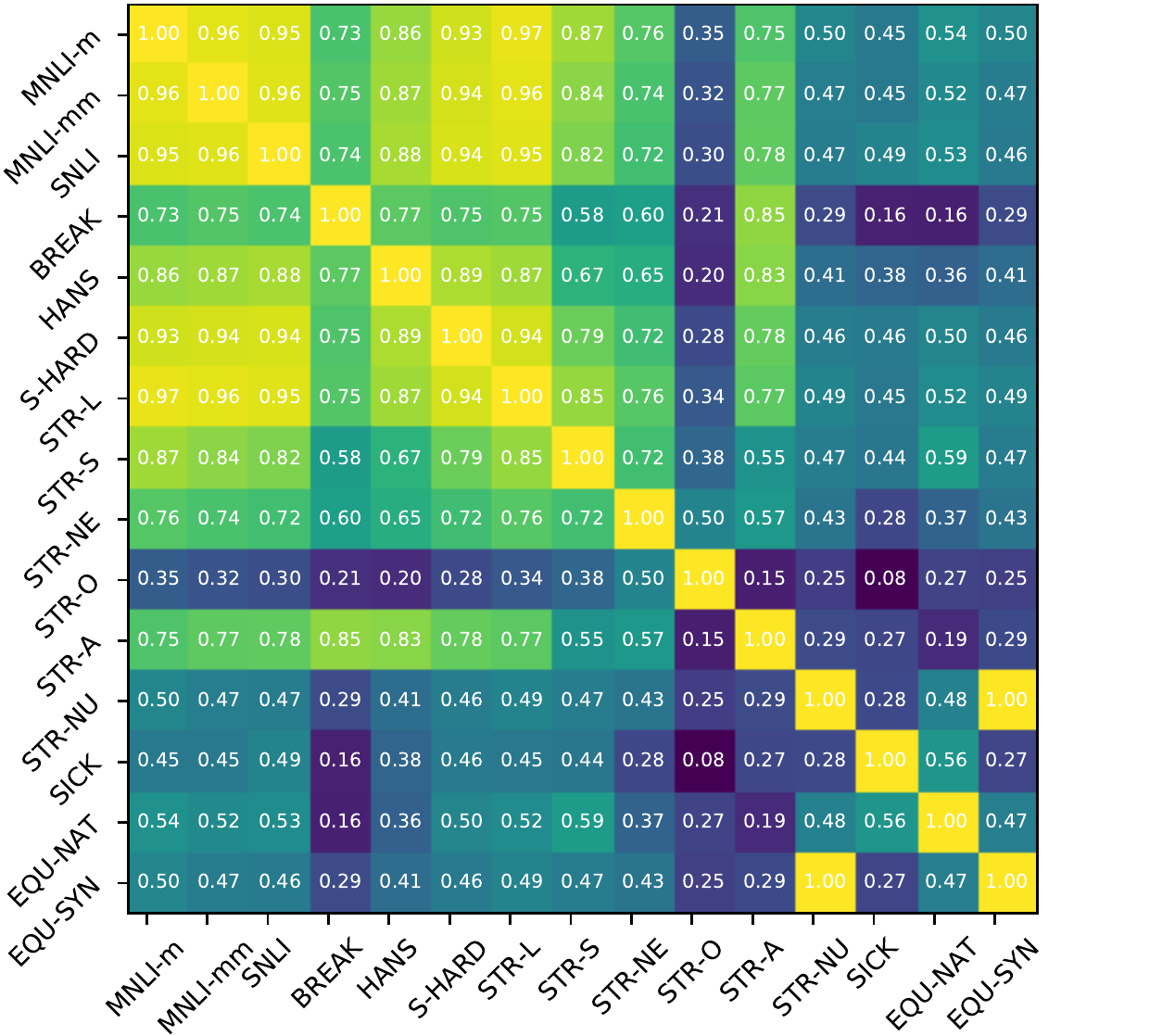}
    \caption{Spearman's correlations for different datasets showing the low correlation between standard datasets (i.e., MNLI-m, MNLI-mm, and SNLI) and all the other analysis datasets.}
    \vspace{-6pt}
    \label{fig:heatmap_modelselection}
\end{figure}

\paragraph{Low Correlation between Datasets.}
The typical routine for neural network model selection requires practitioners to choose the model or checkpoint hinged on the observation of models' performance on the validation set. The routine was followed in all previous NLI analysis studies where models were chosen by the performance on standard validation set and tested on analysis sets. An important assumption behind this routine is that the performance on the validation set should be correlated with the models' general ability. However, as shown in Fig.~\ref{fig:bert_training_trajectory}, the striking difference between the wildly fluctuated training curves for analysis sets and the smooth curves for the standard validation set questions the validity of this assumption.

Therefore, to check the effectiveness of model selection under these instabilities, we checked the correlation for the performance on different datasets during training.
For dataset $\mathcal{D}^i$, we use $a^i_{t,s}$ to denote the accuracy of the checkpoint at $t$-th time step and trained with the seed $s\in S$, where $S$ is the set of all seeds.
We calculate the correlation $\mathrm{Corr}_{i,j}$ between datasets $\mathcal{D}^i$ and  $\mathcal{D}^j$ by:
\begin{align*} 
\mathrm{Corr}_{i,j} \!=\! \frac{1}{\vert S \vert}\sum_{s\in S}\mathrm{Spearman}\left[ (a^i_{t,s})_{t=1}^{T}, 
(a^j_{t,s})_{t=1}^{T} \right]
\end{align*}
where $T$ is the number of checkpoints.

The correlations between different NLI datasets are shown in Fig.~\ref{fig:heatmap_modelselection}. 
We can observe high correlation ($>0.95$) among standard validation datasets (e.g. MNLI-m, MNLI-mm, SNLI) but low correlations between other dataset pairs, especially when pairing STR-O or STR-NU with MNLI or SNLI.
While these low correlations between standard evaluation sets and analysis sets can bring useful insights for analysis, 
this also indicates that: 1) performance on the standard validation set is not representative enough for certain analysis set performances; 2) comparison/conclusions drawn from analysis datasets' results from model selection on standard evaluation sets may be unreliable.

\section{Tracking Instability}
\label{sec:why}

Before answering the question how to handle these instabilities, we first seek the source of the instability to get a better understanding of the issue.
We start with the intuition that high variance could be the result of high inter-example correlation within the dataset, and then provide hints from experimental observations. Next, we show theoretical evidence to  formalize our claim. Finally, we conclude that the major source of variance is the inter-example correlations based on empirical results.

\subsection{Inter-Example Correlations}
Presumably, the wild fluctuation in the training trajectory on different datasets might come from two potential sources. Firstly, the individual prediction of each example may be highly unstable so that the prediction is constantly changing. Secondly, there might be strong inter-example correlations in the datasets such that a large proportion of predictions are more likely to change simultaneously, thus causing large instability.
Here we show that the second reason, i.e., the strong inter-example prediction correlation is the major factor.

\begin{figure}[t]
    \centering
    \includegraphics[width=0.98\linewidth]{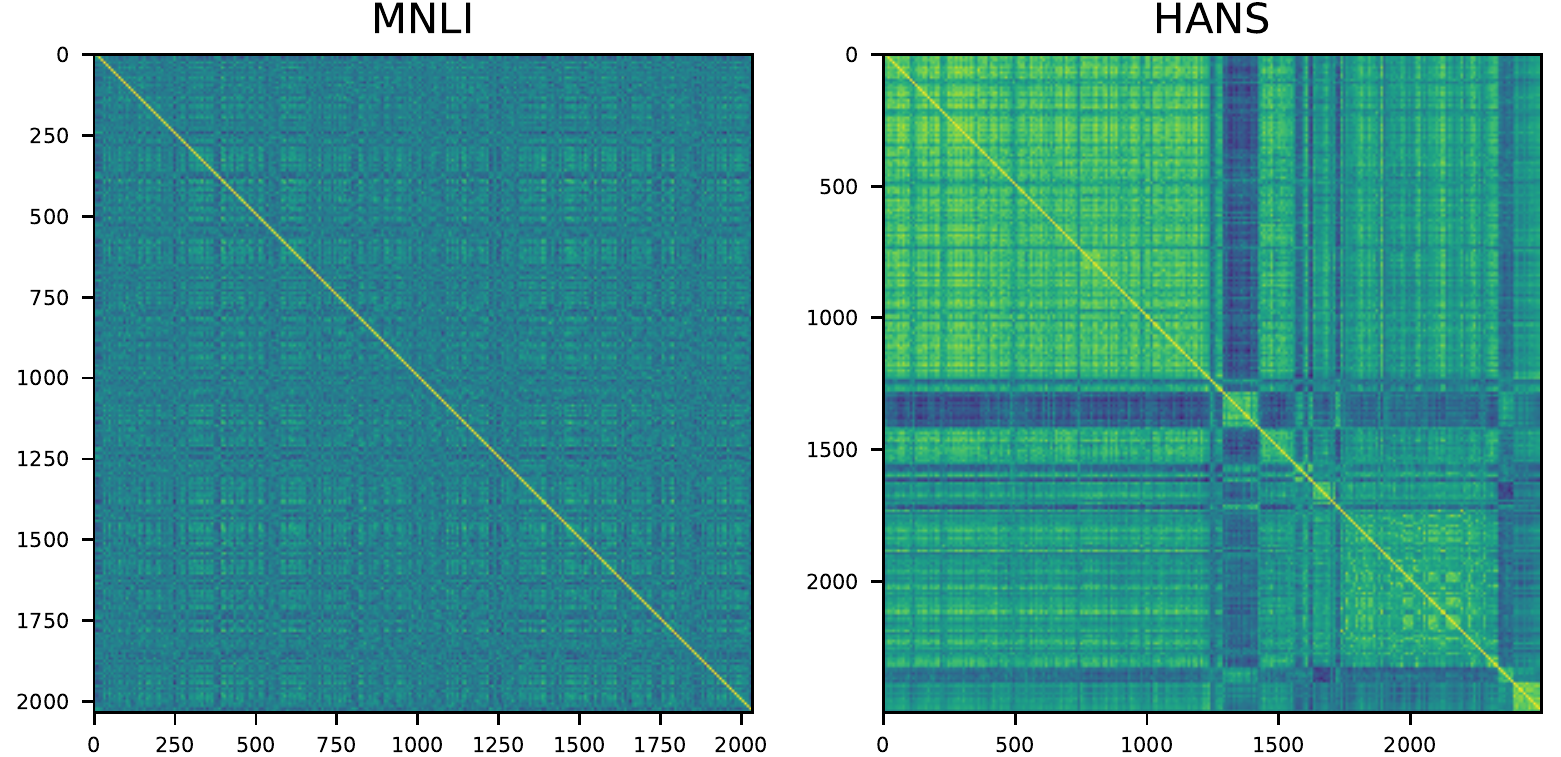}
    \caption{The two heatmaps of inter-example correlations matrices for both MNLI and HANS. Each point in the heatmap represents the Spearman's correlation between the predictions of an example-pair.}
    \vspace{-5pt}
    \label{fig:interexample_map}
    \vspace{-5pt}
\end{figure}

We examine the correlation between different example prediction pairs during the training process.
In Fig.~\ref{fig:interexample_map}, we calculated the inter-example Spearman's correlation on MNLI and HANS.
Fig.~\ref{fig:interexample_map} shows a clear difference between the inter-example correlation in stable (MNLI) datasets versus unstable (HANS) datasets. 
For stable datasets (MNLI), the correlations between the predictions of examples are uniformly low, while for unstable datasets (HANS), there exist clear groups of examples with very strong inter-correlation between their predictions. 
This observation suggests that those groups could be a major source of instability if they contain samples with frequently changing predictions.

\begin{table*}[t]
\resizebox{\textwidth}{!}{%
\begin{tabular}{lccccccccccccccc}
\toprule
\multirow{2}{*}{Statistics}   & \multicolumn{3}{c}{Standard Dataset}                                                  & \multicolumn{12}{c}{Analysis Dataset}    \\ 
\cmidrule(lr){2-4}  \cmidrule(lr){5-16} 
         &\bf   MNLI-m &\bf  MNLI-mm &\bf SNLI   &\bf BREAK &\bf HANS  &\bf  SNLI-hard &\bf  STR-L &\bf  STR-S &\bf  STR-NE &\bf  STR-O &\bf  STR-A &\bf  STR-NU &\bf SICK &\bf EQU-NAT &\bf EQU-SYN \\
\midrule
$\sqrt{\text{Total Var}}$   &    0.24  &    0.20      & 0.11    &  0.38     & 1.51          & 0.40        &  0.34     &   0.28   &   0.65    &   0.90    &   0.89    &   3.76 & 0.35 & 0.66 & 3.47   \\
$\sqrt{\text{Idp Var}}$   &  0.18    & 0.18         &  0.13    &   0.12    &  0.10       & 0.30        &  0.17     &    0.22   &    0.17   &    0.19   &  0.56     &    0.33  & 0.17 & 0.59 & 0.31 \\
$\sqrt{\vert\text{Cov}\vert}$   &  0.16  &  0.09     & 0.06    & 0.36 & 1.51       & 0.27     & 0.28        & 0.15        & 0.63      & 0.88     & 0.69     &  3.74     & 0.31 & 0.31 & 3.45 \\
\bottomrule
\end{tabular}%
}
\caption{
The square roots of total variance (Total Var), independent variance (Idp Var), and the absolute covariance ($\vert\text{Cov}\vert$) of BERT model on different NLI datasets.
Square root is applied
to map variances and covariances to a normal range.
Analysis datasets have much higher covariance than standard datasets. 
}
\vspace{-6pt}
\label{tab:idp_variance_new}
\end{table*}

\begin{table}[t]
\resizebox{0.48\textwidth}{!}{%
\begin{tabular}{lccc}
\toprule
\multirow{2}{*}{Statistics}   & \multicolumn{1}{c}{Standard Dataset}                                                  & \multicolumn{2}{c}{Analysis Dataset}    \\ 
\cmidrule(lr){2-2}  \cmidrule(lr){3-4} 
         &\bf   SQuAD &\bf  AddSent & \bf AddOneSent \\
\midrule
$\sqrt\text{Total Var}$    &    0.13  &    0.57      & 0.48   \\
$\sqrt\text{Idp Var}$   &  0.15    & 0.33         &  0.44    \\
$\sqrt{\vert\text{Cov}\vert}$   & 0.09     &   0.43       &  0.13    \\
\bottomrule
\end{tabular}%
}
\caption{
The square roots of total variance (Total Var), independent variance (Idp Var), and absolute covariance ($\vert\text{Cov}\vert$) of BERT model on different RC datasets. 
}
\label{tab:idp_variance_mc}
\end{table}

\subsection{Variance Decomposition}
\label{sec:v_dec}
Next, we provide theoretical support to show how the high inter-example correlation contributes to the large variance in final accuracy. Later, we will also demonstrate that it is the major source of the large variance.
Suppose dataset $\mathcal{D}$ contains examples $\{x_i, y_i\}_{i=1}^N$, where $N$ is the number of data points in the dataset, $x_i$ and $y_i$ are the inputs and labels, respectively.
We use a random variable $C_{i}$ to denote whether model $M$ predicts the $i$-th example correctly:
$C_i\!=\!\mathbbm{1}[y_i = M(x_i)]$.
We ignore the model symbol $M$ in our later notations for simplicity.
The accuracy $\mathit{Acc}$ of model $M$ is another random variable, which equals to the average over $\{C_i\}$, w.r.t. different model weights (i.e., caused by different random seeds in our experiments):
$\mathit{Acc} = \frac{1}{N} \sum_i C_i$.
We then decompose the variance of the accuracy $\mathrm{Var}(\mathit{Acc})$ into the sum of data variances $\mathrm{Var}(C_i)$, and the sum of inter-data covariances $\mathrm{Cov}(C_i, C_j)$:
\begin{align}
\label{eq:var_dec}
\mathrm{Var}(\mathit{Acc})
&\!=\! \frac{1}{N^2}\mathrm{Cov}\left(\sum_{i=1}^N C_i,\,\, \sum_{j=1}^N C_j\right) \nonumber \\
&\!=\! \frac{1}{N^2}\sum_{i=1}^N \sum_{j=1}^N \mathrm{Cov}\left( C_i,\,\, C_j\right) \nonumber \\
&\!=\!\frac{1}{N^2}\!\sum_{i=1}^N\! \mathrm{Var}(C_i) \!+\! \frac{2}{N^2}\!\sum_{i< j}\!\mathrm{Cov}( C_i\mbox{,}C_j) 
\end{align}
Here, the first term $\frac{1}{N^2}\sum \mathrm{Var}(C_{i})$ means the instability caused by the randomness in individual example prediction and the second term $\frac{2}{N^2}\sum_{i<j}\mathrm{Cov}(C_{i},C_{j})$ means the instability caused by the covariance of the prediction between different examples. The latter covariance term is highly related to the inter-example correlation.

Finally, to demonstrate that the inter-example correlation is the major source of high variance, we calculate the total variance, the independent variance (the 1st term in Eq. \ref{eq:var_dec}), and the covariance (the 2nd term in Eq. \ref{eq:var_dec}) on every dataset in Table \ref{tab:idp_variance_new}. 
In contrast to similar averages of the independent variance on standard and analysis datasets, we found a large gap between the averages of covariances on different datasets.
This different trend of total variance and independent variance proves that the inter-example correlation is the major reason for the difference of variance on the analysis datasets.

\begin{table}[t]
\centering
\small
\begin{tabular}{rl}
\toprule
Premise: & Though the author encouraged the lawyer, \\
& the tourist waited.\\
Hypothesis: & The author encouraged the lawyer.  \\
Label: &  entailment\\
\midrule
Premise: & The lawyer thought that the senators \\
&supported the manager.         \\
Hypothesis: & The senators supported the manager. \\
Label: &  non-entailment\\
\bottomrule
\end{tabular}
\caption{A highly-correlated example pair in the HANS dataset with the BERT model. This example pair have the largest covariance (0.278) among all the pairs.}
\vspace{-3pt}
\label{tab:nli_example}
\end{table}

\begin{table*}[t]
\resizebox{\textwidth}{!}{%
\begin{tabular}{lccccccccccccc}
\toprule
       \bf Target Eval Set  &  \bf MNLI-m  & \bf BREAK & \bf HANS & \bf SNLI-hard & \bf STR-L & \bf STR-S & \bf STR-NE & \bf STR-O & \bf STR-A & \bf STR-NU & \bf SICK & \bf EQU-NAT & \bf EQU-SYN\\
\midrule
\multicolumn{14}{c}{\textit{Accuracy Mean}}\\
\midrule
MNLI-m    &    85.1     &  95.3     & 61.6     & 80.9 &  \textbf{81.9 }       &  77.3  &   55.5   &   59.9   &   62.9   &   41.1 & 57.3 & 60.1 & 41.3       \\
Re-Split Dev   &  -      &   \textbf{96.2}   & \textbf{ 64.3}   & \textbf{81.0 }   &  81.7    &    \textbf{77.4}  &    \textbf{56.5}  &    \textbf{66.0}  & \textbf{ 67.2}     &   \textbf{ 48.2} & \textbf{59.3} & \textbf{61.2} & \textbf{47.6}   \\
\midrule
\multicolumn{14}{c}{\textit{Accuracy Standard Deviation}}\\
\midrule
MNLI-m    &    0.22  &   0.37     & 1.57     &   \textbf{0.33 }       &  0.36     &   \textbf{0.35}  &   \textbf{0.65 }   &   \textbf{0.88}    &   \textbf{1.60}    &   3.49   & \textbf{0.55} & \textbf{1.06} & 3.19    \\
Re-Split Dev   &  -      &   \textbf{0.32}   &  \textbf{1.51}    &     0.52        &  \textbf{0.34}     &    0.47   &    0.83   &    2.70   &  1.83     &    \textbf{2.64}  & 1.26 & 1.18 & \textbf{1.86}    \\
\bottomrule
\end{tabular}%
}
\caption{The comparison of means and standard deviations of the accuracies when model selection are conducted based on different development set. `MNLI-m' chooses the best checkpoint based on the MNLI-m validation set. `Re-Split Dev' chooses the best checkpoint based on the corresponding re-splitted analysis-dev set.}
\vspace{-5pt}
\label{tab:resplit}
\end{table*}

\subsection{Highly-Correlated Cases}
\label{sec:cases}
From these analyses, we can see that one major reason behind the high variance in certain analysis datasets is high inter-example correlation. Following this direction, the next question is why these highly-correlated example-pairs are more likely to appear in analysis datasets. From Table~\ref{tab:all_instability}, we can find that the largest variance happens in HANS, several subsets of STR, and EQU-SYN. On the other hand, while datasets like SNLI-hard and EQU-NAT are also analysis datasets, their variance is much smaller than the former ones. One crucial difference among the high-variance datasets is that they are usually created with the help of synthetic rules. 

This way of well-controlled synthetic-rule based construction can effectively target certain linguistic phenomena in the dataset, but they may also cause many examples to share similar lexicon usage. One example from the HANS dataset is shown in Table \ref{tab:nli_example}, and another similar example for RC is also shown in Appendix. These similarities in syntax and lexicon are very likely to cause the prediction in these two examples to be highly-correlated.
Another evidence can also be seen from Figure~\ref{fig:interexample_map}, where we can see clear boundaries of blocks of high-correlation examples in the right sub-figure (for HANS dataset). Since the examples in HANS are ordered by its templates, examples in the same block are created using the same template. Hence, the block patterns in the figure also show how synthetic rules may cause predictions to be more correlated with each other.

In conclusion, since analysis datasets are sometimes created using pre-specified linguistic patterns/properties and investigation phenomena in mind, the distributions of analysis datasets are less diverse than the distributions of standard datasets. The difficulty of the dataset and the lack of diversity can lead to highly-correlated predictions and high instability in models' final performances.

\section{Implications, Suggestions, and Discussion}
So far, we have demonstrated how severe this instability issue is and how the instability can be traced back to the high correlation between predictions of certain example clusters. Now based on all the previous analysis results, we discuss potential ways of how to deal with this instability issue.

We first want to point out that this instability issue is not a simple problem that can be solved by trivial modifications of the dataset, model, or training algorithm. Here, below we first present one initial attempt at illustrating the difficulty of solving this issue via dataset resplitting.

\paragraph{Limitation of Model Selection.}
In this experiment, we see if an oracle model selection process can help reduce instability.
Unlike the benchmark datasets, such as SNLI, MNLI, and SQuAD, analysis sets are often proposed as a single set without dev/test splits. In Sec.~\ref{sec:why},  we observe that models' performances on analysis sets have little correlation with model performance on standard validation sets, making the selection model routine useless for reducing performance instability on analysis sets.
Therefore, we do oracle  model selection by dividing the original analysis set into an 80\% analysis-dev dataset and a 20\% analysis-test dataset. Model selection is a procedure used to select the best model based on the high correlation between dev/test sets. Hence, the dev/test split here will naturally be expected to have the best performance. 

In Table~\ref{tab:resplit}, we compare the results of BERT-B on the new analysis-test with model selection based on the results on either MNLI or the corresponding analysis-dev. While model selection on analysis-dev helps increase the mean performance on several datasets\footnote{Although the new selection increase the performance mean, we suggest not use the results on analysis sets as benchmark scores but only as toolkits to probe model/architecture changes since analysis datasets are easy to overfit.}, especially on HANS, STR-O, and STR-NU, indicating the expected high correlation inside the analysis set, however, the variances of final results are not always reduced for different datasets. Hence, besides the performance instability caused by noisy model selection, different random seeds indeed lead to models with different performance on analysis datasets. This observation might indicate that performance instability is relatively independent of the mean performance and hints that current models may have intrinsic randomness brought by different random seeds which is unlikely to be removed through simple dataset/model fixes.

\subsection{Implications of Result Instability}
If the intrinsic randomness in the model prevents a quick fix, what does this instability issue imply?
At first glance, one may view the instability as a problem caused by careless dataset design or deficiency in model architecture/training algorithms. While both parts are indeed imperfect, here we suggest it is more beneficial to view this instability as an inevitable consequence of the current datasets and models. On the data side, as these analysis datasets usually leverage specific rules or linguistic patterns to generate examples targeting specific linguistic phenomena and properties, they contain highly similar examples (examples shown in \ref{sec:cases}). Hence, the model's predictions of these examples will be inevitably highly-correlated. On the model side, as the current model is not good enough to stably capture these hard linguistic/logical properties through learning, they will exhibit instability over some examples, which is amplified by the high correlation between examples' predictions. These datasets can still serve as good evaluation tools as long as we are aware of the instability issue and report results with multiple runs. To better handle the instability, we also propose some long and short term solution suggestions below, based on variance reporting and analysis dataset diversification.

\subsection{Short/Long Term Suggestions}

\paragraph{Better Analysis Reporting (Short Term).} 
Even if we cannot get a quick fix to remove the instability in the results, it is still important to keep making progress using currently available resources, and more importantly, to accurately evaluate this progress.
Therefore, in the short run, we encourage researchers to report the decomposed variance (Idp Var and Cov) for a more accurate understanding of the models and datasets as in Sec \ref{sec:v_dec}, Table \ref{tab:idp_variance_new} and Table \ref{tab:idp_variance_mc}.
The first number (independent variance, i.e., Idp Var) can be viewed as a metric regarding how stable the model makes one single prediction and this number can be compared across different models.
Models with a lower score can be interpreted as being more stable for one single prediction. The values of Cov also help us better understand both the model and the datasets. A high Cov indicates that many examples look similar to the model, and the model may be exploiting some common artifacts in this group of examples. A lower Cov usually means that the dataset is diverse and is preferable for evaluation.
By comparing models with both total variance and the Idp Var, we can have a better understanding of where the instability of the models comes from.
A more stable model should aim to improve the total variance with more focus on Idp Var. If the target is to learn the targeted property of the dataset better, then more focus should be drawn towards the second term when analysing the results. 

\paragraph{Model and Dataset Suggestions (Long Term).}
In the long run, we should be focusing on improving models (including better inductive biases, large-scale pre-training with tasks concerning structure/compositionality) so that they can get high accuracy stably. 
Dataset-wise, we encourage the construction of more diverse datasets (in terms of syntax and lexicon). From our previous results and analysis in Section~\ref{sec:why}, we can see that analysis datasets from natural real-life sources usually lead to lower covariance between predictions and show better stability. Manual verification for synthetic examples also helps reduce the instability of analysis datasets. While controlled synthetic datasets are more accurate and effective in evaluating certain linguistic phenomenon, the lack of diversity may increase the model's ability to guess the answer right and solve only that single pattern/property instead of mastering the systematic capability of those linguistic properties under different contexts (as reflected by the poor correlation between different analysis datasets). Therefore, a very valuable direction in constructing these datasets is to both maintain the specificity of the dataset while having a larger diversity.

\section{Conclusions}
Auxiliary analysis datasets are meant to be important resources for debugging and understanding models. However, large instability of current models on some of these analysis sets undermine such benefits and bring non-ignorable obstacles for future research. In this paper, we examine the issue of instability in detail, provide theoretical and empirical evidence discovering the high inter-example correlation that causes this issue. Finally, we give suggestions on future research directions and on better analysis variance reporting. We hope this paper will guide researchers on how to handle instability and inspire future work in this direction.

\section*{Acknowledgments}
We thank the reviewers for their helpful comments. This work was supported by ONR Grant N00014-18-1-2871, DARPA YFA17-D17AP00022, and NSF-CAREER Award 1846185. The views contained in this article are those of the authors and not of the funding agency.

\bibliography{acl2020}

\begin{thebibliography}{56}
\expandafter\ifx\csname natexlab\endcsname\relax\def\natexlab#1{#1}\fi

\bibitem[{Bloodgood and Grothendieck(2013)}]{bloodgood2013analysis}
Michael Bloodgood and John Grothendieck. 2013.
\newblock Analysis of stopping active learning based on stabilizing
  predictions.
\newblock In \emph{Proceedings of the Seventeenth Conference on Computational
  Natural Language Learning}, pages 10--19.

\bibitem[{Bowman et~al.(2015)Bowman, Angeli, Potts, and
  Manning}]{bowman2015large}
Samuel~R Bowman, Gabor Angeli, Christopher Potts, and Christopher~D Manning.
  2015.
\newblock A large annotated corpus for learning natural language inference.
\newblock In \emph{Proceedings of the 2015 Conference on Empirical Methods in
  Natural Language Processing}, pages 632--642.

\bibitem[{Carmona et~al.(2018)Carmona, Mitchell, and
  Riedel}]{carmona2018behavior}
Vicente Iv{\'a}n~S{\'a}nchez Carmona, Jeff Mitchell, and Sebastian Riedel.
  2018.
\newblock Behavior analysis of nli models: Uncovering the influence of three
  factors on robustness.
\newblock \emph{NAACL}.

\bibitem[{Chen et~al.(2017)Chen, Zhu, Ling, Wei, Jiang, and
  Inkpen}]{chen2017enhanced}
Qian Chen, Xiaodan Zhu, Zhen-Hua Ling, Si~Wei, Hui Jiang, and Diana Inkpen.
  2017.
\newblock Enhanced lstm for natural language inference.
\newblock In \emph{Proceedings of the 55th Annual Meeting of the Association
  for Computational Linguistics (Volume 1: Long Papers)}, pages 1657--1668.

\bibitem[{Clark et~al.(2019)Clark, Yatskar, and Zettlemoyer}]{clark2019don}
Christopher Clark, Mark Yatskar, and Luke Zettlemoyer. 2019.
\newblock Don’t take the easy way out: Ensemble based methods for avoiding
  known dataset biases.
\newblock In \emph{Proceedings of the 2019 Conference on Empirical Methods in
  Natural Language Processing and the 9th International Joint Conference on
  Natural Language Processing (EMNLP-IJCNLP)}, pages 4060--4073.

\bibitem[{Dai et~al.(2019)Dai, Yang, Yang, Carbonell, Le, and
  Salakhutdinov}]{dai2019transformer}
Zihang Dai, Zhilin Yang, Yiming Yang, Jaime~G Carbonell, Quoc Le, and Ruslan
  Salakhutdinov. 2019.
\newblock Transformer-xl: Attentive language models beyond a fixed-length
  context.
\newblock In \emph{Proceedings of the 57th Annual Meeting of the Association
  for Computational Linguistics}, pages 2978--2988.

\bibitem[{Devlin et~al.(2019)Devlin, Chang, Lee, and
  Toutanova}]{devlin2019bert}
Jacob Devlin, Ming-Wei Chang, Kenton Lee, and Kristina Toutanova. 2019.
\newblock Bert: Pre-training of deep bidirectional transformers for language
  understanding.
\newblock In \emph{Proceedings of the 2019 Conference of the North American
  Chapter of the Association for Computational Linguistics: Human Language
  Technologies, Volume 1 (Long and Short Papers)}, pages 4171--4186.

\bibitem[{Dodge et~al.(2020)Dodge, Ilharco, Schwartz, Farhadi, Hajishirzi, and
  Smith}]{dodge2020fine}
Jesse Dodge, Gabriel Ilharco, Roy Schwartz, Ali Farhadi, Hannaneh Hajishirzi,
  and Noah Smith. 2020.
\newblock Fine-tuning pretrained language models: Weight initializations, data
  orders, and early stopping.
\newblock \emph{arXiv preprint arXiv:2002.06305}.

\bibitem[{Gardner et~al.(2018)Gardner, Grus, Neumann, Tafjord, Dasigi, Liu,
  Peters, Schmitz, and Zettlemoyer}]{Gardner2017AllenNLP}
Matt Gardner, Joel Grus, Mark Neumann, Oyvind Tafjord, Pradeep Dasigi, Nelson~F
  Liu, Matthew Peters, Michael Schmitz, and Luke Zettlemoyer. 2018.
\newblock Allennlp: A deep semantic natural language processing platform.
\newblock In \emph{Proceedings of Workshop for NLP Open Source Software
  (NLP-OSS)}, pages 1--6.

\bibitem[{Geiger et~al.(2019)Geiger, Cases, Karttunen, and
  Potts}]{geiger2019posing}
Atticus Geiger, Ignacio Cases, Lauri Karttunen, and Chris Potts. 2019.
\newblock Posing fair generalization tasks for natural language inference.
\newblock \emph{EMNLP}.

\bibitem[{Geva et~al.(2019)Geva, Goldberg, and Berant}]{geva2019we}
Mor Geva, Yoav Goldberg, and Jonathan Berant. 2019.
\newblock Are we modeling the task or the annotator? an investigation of
  annotator bias in natural language understanding datasets.
\newblock \emph{EMNLP}.

\bibitem[{Glockner et~al.(2018)Glockner, Shwartz, and
  Goldberg}]{glockner2018breaking}
Max Glockner, Vered Shwartz, and Yoav Goldberg. 2018.
\newblock Breaking nli systems with sentences that require simple lexical
  inferences.
\newblock In \emph{Proceedings of the 56th Annual Meeting of the Association
  for Computational Linguistics (Volume 2: Short Papers)}, pages 650--655.

\bibitem[{Goyal et~al.(2017)Goyal, Khot, Summers-Stay, Batra, and
  Parikh}]{goyal2017making}
Yash Goyal, Tejas Khot, Douglas Summers-Stay, Dhruv Batra, and Devi Parikh.
  2017.
\newblock Making the v in vqa matter: Elevating the role of image understanding
  in visual question answering.
\newblock In \emph{Proceedings of the IEEE Conference on Computer Vision and
  Pattern Recognition}, pages 6904--6913.

\bibitem[{Gururangan et~al.(2018)Gururangan, Swayamdipta, Levy, Schwartz,
  Bowman, and Smith}]{gururangan2018annotation}
Suchin Gururangan, Swabha Swayamdipta, Omer Levy, Roy Schwartz, Samuel Bowman,
  and Noah~A Smith. 2018.
\newblock Annotation artifacts in natural language inference data.
\newblock In \emph{Proceedings of the 2018 Conference of the North American
  Chapter of the Association for Computational Linguistics: Human Language
  Technologies, Volume 2 (Short Papers)}, pages 107--112.

\bibitem[{He et~al.(2019)He, Zha, and Wang}]{he2019unlearn}
He~He, Sheng Zha, and Haohan Wang. 2019.
\newblock Unlearn dataset bias in natural language inference by fitting the
  residual.
\newblock In \emph{Proceedings of the 2nd Workshop on Deep Learning Approaches
  for Low-Resource NLP (DeepLo 2019)}, pages 132--142.

\bibitem[{Irpan(2018)}]{rlblogpost}
Alex Irpan. 2018.
\newblock Deep reinforcement learning doesn't work yet.
\newblock \url{https://www.alexirpan.com/2018/02/14/rl-hard.html}.

\bibitem[{Jabri et~al.(2016)Jabri, Joulin, and Van
  Der~Maaten}]{jabri2016revisiting}
Allan Jabri, Armand Joulin, and Laurens Van Der~Maaten. 2016.
\newblock Revisiting visual question answering baselines.
\newblock In \emph{European conference on computer vision}, pages 727--739.
  Springer.

\bibitem[{Jeretic et~al.(2020)Jeretic, Warstadt, Bhooshan, and
  Williams}]{jeretic-etal-2020-natural}
Paloma Jeretic, Alex Warstadt, Suvrat Bhooshan, and Adina Williams. 2020.
\newblock \href {https://doi.org/10.18653/v1/2020.acl-main.768} {Are natural
  language inference models {IMPPRESsive}? {L}earning {IMPlicature} and
  {PRESupposition}}.
\newblock In \emph{Proceedings of the 58th Annual Meeting of the Association
  for Computational Linguistics}, Online. Association for Computational
  Linguistics.

\bibitem[{Jia and Liang(2017)}]{jia2017advsquad}
Robin Jia and Percy Liang. 2017.
\newblock Adversarial examples for evaluating reading comprehension systems.
\newblock \emph{EMNLP}.

\bibitem[{Kaushik and Lipton(2018)}]{kaushik2018much}
Divyansh Kaushik and Zachary~C Lipton. 2018.
\newblock How much reading does reading comprehension require? a critical
  investigation of popular benchmarks.
\newblock \emph{EMNLP}.

\bibitem[{Lewis and Fan(2019)}]{Lewis2019GenerativeQA}
Mike Lewis and Angela Fan. 2019.
\newblock Generative question answering: Learning to answer the whole question.
\newblock In \emph{ICLR}.

\bibitem[{Liu et~al.(2019{\natexlab{a}})Liu, Schwartz, and
  Smith}]{liu2019inoculation}
Nelson~F Liu, Roy Schwartz, and Noah~A Smith. 2019{\natexlab{a}}.
\newblock Inoculation by fine-tuning: A method for analyzing challenge
  datasets.
\newblock \emph{NAACL}.

\bibitem[{Liu et~al.(2019{\natexlab{b}})Liu, Ott, Goyal, Du, Joshi, Chen, Levy,
  Lewis, Zettlemoyer, and Stoyanov}]{liu2019roberta}
Yinhan Liu, Myle Ott, Naman Goyal, Jingfei Du, Mandar Joshi, Danqi Chen, Omer
  Levy, Mike Lewis, Luke Zettlemoyer, and Veselin Stoyanov. 2019{\natexlab{b}}.
\newblock Roberta: A robustly optimized bert pretraining approach.
\newblock \emph{arXiv preprint arXiv:1907.11692}.

\bibitem[{Makel et~al.(2012)Makel, Plucker, and
  Hegarty}]{makel2012replications}
Matthew~C Makel, Jonathan~A Plucker, and Boyd Hegarty. 2012.
\newblock Replications in psychology research: How often do they really occur?
\newblock \emph{Perspectives on Psychological Science}, 7(6):537--542.

\bibitem[{Marelli et~al.(2014)Marelli, Menini, Baroni, Bentivogli, Bernardi,
  Zamparelli et~al.}]{marelli2014sick}
Marco Marelli, Stefano Menini, Marco Baroni, Luisa Bentivogli, Raffaella
  Bernardi, Roberto Zamparelli, et~al. 2014.
\newblock A sick cure for the evaluation of compositional distributional
  semantic models.
\newblock In \emph{LREC}.

\bibitem[{McCoy et~al.(2019{\natexlab{a}})McCoy, Min, and
  Linzen}]{mccoy2019berts}
R~Thomas McCoy, Junghyun Min, and Tal Linzen. 2019{\natexlab{a}}.
\newblock Berts of a feather do not generalize together: Large variability in
  generalization across models with similar test set performance.
\newblock \emph{arXiv preprint arXiv:1911.02969}.

\bibitem[{McCoy et~al.(2019{\natexlab{b}})McCoy, Pavlick, and
  Linzen}]{mccoy2019right}
Tom McCoy, Ellie Pavlick, and Tal Linzen. 2019{\natexlab{b}}.
\newblock Right for the wrong reasons: Diagnosing syntactic heuristics in
  natural language inference.
\newblock In \emph{Proceedings of the 57th Annual Meeting of the Association
  for Computational Linguistics}, pages 3428--3448.

\bibitem[{Minervini and Riedel(2018)}]{minervini2018adversarially}
Pasquale Minervini and Sebastian Riedel. 2018.
\newblock Adversarially regularising neural nli models to integrate logical
  background knowledge.
\newblock \emph{CoNLL}.

\bibitem[{Moosavi et~al.(2019)Moosavi, Utama, R{\"u}ckl{\'e}, and
  Gurevych}]{moosavi2019improving}
Nafise~Sadat Moosavi, Prasetya~Ajie Utama, Andreas R{\"u}ckl{\'e}, and Iryna
  Gurevych. 2019.
\newblock Improving generalization by incorporating coverage in natural
  language inference.
\newblock \emph{arXiv preprint arXiv:1909.08940}.

\bibitem[{Naik et~al.(2018{\natexlab{a}})Naik, Ravichander, Sadeh, Rose, and
  Neubig}]{naik2018stress}
Aakanksha Naik, Abhilasha Ravichander, Norman Sadeh, Carolyn Rose, and Graham
  Neubig. 2018{\natexlab{a}}.
\newblock Stress test evaluation for natural language inference.
\newblock In \emph{Proceedings of the 27th International Conference on
  Computational Linguistics}, pages 2340--2353.

\bibitem[{Naik et~al.(2018{\natexlab{b}})Naik, Ravichander, Sadeh, Rose, and
  Neubig}]{naik-etal-2018-stress}
Aakanksha Naik, Abhilasha Ravichander, Norman Sadeh, Carolyn Rose, and Graham
  Neubig. 2018{\natexlab{b}}.
\newblock \href {https://www.aclweb.org/anthology/C18-1198} {Stress test
  evaluation for natural language inference}.
\newblock In \emph{Proceedings of the 27th International Conference on
  Computational Linguistics}, pages 2340--2353, Santa Fe, New Mexico, USA.
  Association for Computational Linguistics.

\bibitem[{Nie et~al.(2020{\natexlab{a}})Nie, Bauer, and
  Bansal}]{nie-etal-2020-simple}
Yixin Nie, Lisa Bauer, and Mohit Bansal. 2020{\natexlab{a}}.
\newblock \href {https://www.aclweb.org/anthology/2020.fever-1.1} {Simple
  compounded-label training for fact extraction and verification}.
\newblock In \emph{Proceedings of the Third Workshop on Fact Extraction and
  VERification (FEVER)}, Online. Association for Computational Linguistics.

\bibitem[{Nie et~al.(2019)Nie, Wang, and Bansal}]{nie2019analyzing}
Yixin Nie, Yicheng Wang, and Mohit Bansal. 2019.
\newblock Analyzing compositionality-sensitivity of nli models.
\newblock In \emph{Proceedings of the AAAI Conference on Artificial
  Intelligence}, volume~33, pages 6867--6874.

\bibitem[{Nie et~al.(2020{\natexlab{b}})Nie, Williams, Dinan, Bansal, Weston,
  and Kiela}]{nie2019adversarial}
Yixin Nie, Adina Williams, Emily Dinan, Mohit Bansal, Jason Weston, and Douwe
  Kiela. 2020{\natexlab{b}}.
\newblock Adversarial nli: A new benchmark for natural language understanding.
\newblock \emph{ACL}.

\bibitem[{Nie et~al.(2020{\natexlab{c}})Nie, Zhou, and
  Bansal}]{ynie2020chaosnli}
Yixin Nie, Xiang Zhou, and Mohit Bansal. 2020{\natexlab{c}}.
\newblock What can we learn from collective human opinions on natural language
  inference data?
\newblock In \emph{Proceedings of the 2020 Conference on Empirical Methods in
  Natural Language Processing (EMNLP)}. Association for Computational
  Linguistics.

\bibitem[{Pang et~al.(2019)Pang, Lin, and Smith}]{pang2019improving}
Deric Pang, Lucy~H Lin, and Noah~A Smith. 2019.
\newblock Improving natural language inference with a pretrained parser.
\newblock \emph{arXiv preprint arXiv:1909.08217}.

\bibitem[{Peters et~al.(2018)Peters, Neumann, Iyyer, Gardner, Clark, Lee, and
  Zettlemoyer}]{Peters_ELMO@2018}
Matthew~E. Peters, Mark Neumann, Mohit Iyyer, Matt Gardner, Christopher Clark,
  Kenton Lee, and Luke Zettlemoyer. 2018.
\newblock Deep contextualized word representations.
\newblock In \emph{NAACL}.

\bibitem[{Phang et~al.(2018)Phang, F{\'e}vry, and Bowman}]{phang2018sentence}
Jason Phang, Thibault F{\'e}vry, and Samuel~R Bowman. 2018.
\newblock Sentence encoders on stilts: Supplementary training on intermediate
  labeled-data tasks.
\newblock \emph{arXiv preprint arXiv:1811.01088}.

\bibitem[{Radford et~al.(2018)Radford, Narasimhan, Salimans, and
  Sutskever}]{radford2018gpt_improving}
Alec Radford, Karthik Narasimhan, Tim Salimans, and Ilya Sutskever. 2018.
\newblock Improving language understanding by generative pre-training.
\newblock \emph{URL https://s3-us-west-2. amazonaws.
  com/openai-assets/researchcovers/languageunsupervised/language understanding
  paper. pdf}.

\bibitem[{Radford et~al.(2019)Radford, Wu, Child, Luan, Amodei, and
  Sutskever}]{radford2019gpt2_language}
Alec Radford, Jeffrey Wu, Rewon Child, David Luan, Dario Amodei, and Ilya
  Sutskever. 2019.
\newblock Language models are unsupervised multitask learners.
\newblock \emph{OpenAI Blog}, 1(8).

\bibitem[{Rajpurkar et~al.(2018)Rajpurkar, Jia, and Liang}]{rajpurkar2018know}
Pranav Rajpurkar, Robin Jia, and Percy Liang. 2018.
\newblock Know what you don't know: Unanswerable questions for squad.
\newblock \emph{ACL}.

\bibitem[{Rajpurkar et~al.(2016{\natexlab{a}})Rajpurkar, Zhang, Lopyrev, and
  Liang}]{rajpurkar-etal-2016-squad}
Pranav Rajpurkar, Jian Zhang, Konstantin Lopyrev, and Percy Liang.
  2016{\natexlab{a}}.
\newblock \href {https://doi.org/10.18653/v1/D16-1264} {{SQ}u{AD}: 100,000+
  questions for machine comprehension of text}.
\newblock In \emph{Proceedings of the 2016 Conference on Empirical Methods in
  Natural Language Processing}, pages 2383--2392, Austin, Texas. Association
  for Computational Linguistics.

\bibitem[{Rajpurkar et~al.(2016{\natexlab{b}})Rajpurkar, Zhang, Lopyrev, and
  Liang}]{rajpurkar2016squad}
Pranav Rajpurkar, Jian Zhang, Konstantin Lopyrev, and Percy Liang.
  2016{\natexlab{b}}.
\newblock Squad: 100,000+ questions for machine comprehension of text.
\newblock \emph{EMNLP}.

\bibitem[{Ravichander et~al.(2019)Ravichander, Naik, Rose, and
  Hovy}]{ravichander2019equate}
Abhilasha Ravichander, Aakanksha Naik, Carolyn Rose, and Eduard Hovy. 2019.
\newblock Equate: A benchmark evaluation framework for quantitative reasoning
  in natural language inference.
\newblock In \emph{Proceedings of the 23rd Conference on Computational Natural
  Language Learning (CoNLL)}, pages 349--361.

\bibitem[{Reimers and Gurevych(2017)}]{reimers2017reporting}
Nils Reimers and Iryna Gurevych. 2017.
\newblock Reporting score distributions makes a difference: Performance study
  of lstm-networks for sequence tagging.
\newblock \emph{EMNLP}.

\bibitem[{Reimers and Gurevych(2018)}]{reimers2018comparing}
Nils Reimers and Iryna Gurevych. 2018.
\newblock Why comparing single performance scores does not allow to draw
  conclusions about machine learning approaches.
\newblock \emph{arXiv preprint arXiv:1803.09578}.

\bibitem[{Rozen et~al.(2019)Rozen, Shwartz, Aharoni, and
  Dagan}]{rozen2019diversify}
Ohad Rozen, Vered Shwartz, Roee Aharoni, and Ido Dagan. 2019.
\newblock Diversify your datasets: Analyzing generalization via controlled
  variance in adversarial datasets.
\newblock \emph{CoNLL}.

\bibitem[{Shwartz and Dagan(2018)}]{shwartz2018paraphrase}
Vered Shwartz and Ido Dagan. 2018.
\newblock Paraphrase to explicate: Revealing implicit noun-compound relations.
\newblock \emph{ACL}.

\bibitem[{Tsuchiya(2018)}]{tsuchiya2018performance}
Masatoshi Tsuchiya. 2018.
\newblock Performance impact caused by hidden bias of training data for
  recognizing textual entailment.
\newblock \emph{LREC}.

\bibitem[{Vaswani et~al.(2017)Vaswani, Shazeer, Parmar, Uszkoreit, Jones,
  Gomez, Kaiser, and Polosukhin}]{vaswani2017attention}
Ashish Vaswani, Noam Shazeer, Niki Parmar, Jakob Uszkoreit, Llion Jones,
  Aidan~N Gomez, {\L}ukasz Kaiser, and Illia Polosukhin. 2017.
\newblock Attention is all you need.
\newblock In \emph{Advances in neural information processing systems}, pages
  5998--6008.

\bibitem[{Wang et~al.(2019)Wang, Singh, Michael, Hill, Levy, and
  Bowman}]{wang2019glue}
Alex Wang, Amanpreet Singh, Julian Michael, Felix Hill, Omer Levy, and
  Samuel~R. Bowman. 2019.
\newblock {GLUE}: A multi-task benchmark and analysis platform for natural
  language understanding.
\newblock In \emph{Proceedings of ICLR}.

\bibitem[{Weber et~al.(2018)Weber, Shekhar, and
  Balasubramanian}]{weber2018fine}
Noah Weber, Leena Shekhar, and Niranjan Balasubramanian. 2018.
\newblock The fine line between linguistic generalization and failure in
  seq2seq-attention models.
\newblock \emph{NAACL}.

\bibitem[{Williams et~al.(2018)Williams, Nangia, and
  Bowman}]{williams2018broad}
Adina Williams, Nikita Nangia, and Samuel Bowman. 2018.
\newblock A broad-coverage challenge corpus for sentence understanding through
  inference.
\newblock In \emph{Proceedings of the 2018 Conference of the North American
  Chapter of the Association for Computational Linguistics: Human Language
  Technologies, Volume 1 (Long Papers)}, pages 1112--1122.

\bibitem[{Yaghoobzadeh et~al.(2019)Yaghoobzadeh, Tachet, Hazen, and
  Sordoni}]{yaghoobzadeh2019robust}
Yadollah Yaghoobzadeh, Remi Tachet, TJ~Hazen, and Alessandro Sordoni. 2019.
\newblock Robust natural language inference models with example forgetting.
\newblock \emph{arXiv preprint arXiv:1911.03861}.

\bibitem[{Yang et~al.(2019)Yang, Dai, Yang, Carbonell, Salakhutdinov, and
  Le}]{yang2019xlnet}
Zhilin Yang, Zihang Dai, Yiming Yang, Jaime Carbonell, Ruslan Salakhutdinov,
  and Quoc~V Le. 2019.
\newblock Xlnet: Generalized autoregressive pretraining for language
  understanding.
\newblock \emph{NeurIPS}.

\bibitem[{Yeh and Chen(2019)}]{yeh2019qainfomax}
Yi-Ting Yeh and Yun-Nung Chen. 2019.
\newblock Qainfomax: Learning robust question answering system by mutual
  information maximization.
\newblock In \emph{Proceedings of the 2019 Conference on Empirical Methods in
  Natural Language Processing and the 9th International Joint Conference on
  Natural Language Processing (EMNLP-IJCNLP)}, pages 3361--3366.

\end{thebibliography}
\bibliographystyle{acl_natbib}

\section*{Appendix}
\appendix

\section{Details of Models}
For models, we mainly focus on the current state-of-the-art models with a pre-trained transformer structure.
In addition, we also selected several traditional models to see how different structures and the use of pre-trained representations influence the result. 
\subsection{Transformer Models}
\paragraph{BERT~\cite{devlin2019bert}.} BERT is a Transformer model pre-trained with masked language supervision on a large unlabeled corpus to obtain deep bi-directional representations  \cite{vaswani2017attention}. To conduct the task of NLI, the premise and the hypothesis are concatenated as the input and a simple classifier is added on top of these pre-trained representations to predict the label. Similarly, for RC, the question and the passage are concatenated as a single input and the start/end location of the answer span is predicted by computing a dot product between the start/end vector and all the words in the document. The whole model is fine-tuned on NLI/RC datasets before evaluation.

\paragraph{RoBERTa~\cite{liu2019roberta}.} RoBERTa uses the same structure as BERT, but carefully tunes the hyper-parameters for pre-training and is trained 10 times more data during pre-training. The fine-tuning architecture and process are the same as BERT.

\paragraph{XLNet~\cite{yang2019xlnet}.} XLNet also adopts the Transformer structure but the pre-training target is a generalized auto-regressive language modeling. It also can take in infinite-length input by using the Transformer-XL~\cite{dai2019transformer} architecture. The fine-tuning architecture and process are the same as BERT.

\subsection{Traditional Models}
\paragraph{ESIM~\cite{chen2017enhanced}.} ESIM first uses BiLSTM to encode both the premise and the hypothesis sentence and perform cross-attention before making the prediction using a classifier. It is one representative model before the use of pre-trained Transformer structure.

\begin{table}[t]
\resizebox{0.47\textwidth}{!}{%
\begin{tabular}{lccc}
\toprule
\bf Name     & \bf Standard/Analysis  & \bf \#Examples & \bf \#Classes  \\ 
\midrule
\textbf{MNLI-m}    & Standard   &   9815  & 3  \\
\textbf{MNLI-mm}    &  Standard  &    9832 & 3   \\
\textbf{SNLI}    & Standard   &    9842  & 3  \\
\textbf{BREAK-NLI}    &  Analysis   &    8193 & 3  \\
\textbf{HANS}    & Analysis   &    30000 & 2    \\
\textbf{SNLI-hard}    &  Analysis   &    3261 & 3   \\
\textbf{STR-L}    & Analysis   &    9815 & 3    \\
\textbf{STR-S}   &  Analysis   &    8243 & 3    \\
\textbf{STR-NE}    & Analysis   &   9815 & 3    \\
\textbf{STR-O}    &  Analysis  &    9815 & 3    \\
\textbf{STR-A}    & Analysis  &    1561 & 3     \\
\textbf{STR-NU}   &  Analysis   &    7596 & 3    \\
\textbf{SICK}     & Analysis & 9841 & 3 \\
\textbf{EQU-NAT}     & Analysis & 1384 & 3 \\
\textbf{EQU-SYN}     & Analysis & 8318 & 3 \\
\bottomrule
\end{tabular}%
}
\caption{Dataset statistics and categories for all the NLI dev/analysis datasets.}
\label{tab:dataset_stats}
\end{table}

\begin{table}[t]
\resizebox{0.47\textwidth}{!}{%
\begin{tabular}{lccc}
\toprule
\bf Name     & \bf Standard/Analysis  & \bf \#Paragraphs & \bf \#Questions  \\ 
\midrule
\textbf{SQuAD}    & Standard   &   48  & 10570  \\
\textbf{AddSent}    &  Analysis  &    48 & 3560   \\
\textbf{AddOneSent}    & Analysis   &    48  & 1787  \\
\bottomrule
\end{tabular}%
}
\caption{Dataset statistics and categories for all the RC dev/analysis datasets.}
\label{tab:dataset_stats_rc}
\end{table}

\begin{table*}[t]
\resizebox{\textwidth}{!}{%
\begin{tabular}{lccccccccccccccc}
\toprule
\multirow{2}{*}{\textbf{Model}}   & \multicolumn{3}{c}{Standard Datasets}                                                  & \multicolumn{12}{c}{Analysis Sets}    \\ 
\cmidrule(lr){2-4}  \cmidrule(lr){5-16} 
         & \bf  MNLI-m & \bf MNLI-mm & \bf SNLI &  \bf BREAK-NLI & \bf HANS  & \bf SNLI-hard & \bf STR-L & \bf STR-S & \bf  STR-NE & \bf STR-O & \bf STR-A & \bf STR-NU &\bf SICK &\bf EQU-NAT &\bf EQU-SYN \\
\midrule
\textbf{ESIM}          & 77.38$\pm$0.32     &  77.03$\pm$0.18        &  88.34$\pm$0.24  &  78.49$\pm$1.00     &  49.89$\pm$0.15          & 75.03$\pm$0.40        &   74.21$\pm$0.24    &  69.30$\pm$2.38     & 51.61$\pm$1.13      &   57.95$\pm$1.47    &   53.21$\pm$2.04    &  21.02$\pm$1.00   &55.55$\pm$0.47   &55.87$\pm$1.01   &22.89$\pm$0.94     \\
\textbf{ESIM+ELMo}     &  79.83$\pm$0.11  &  79.85$\pm$0.21        &  88.81$\pm$0.17  &    83.24$\pm$1.33   &    50.07$\pm$0.27        &   76.30$\pm$0.45      &  76.29$\pm$0.33     &  74.03$\pm$0.25     &  52.80$\pm$0.79     & 58.42$\pm$1.63     &  54.41$\pm$1.69     &   20.95$\pm$1.00    &57.21$\pm$0.25   &59.19$\pm$0.69   &22.70$\pm$1.01    \\
\textbf{BERT-B}     &    84.72 $\pm$0.24 &    84.89 $\pm$0.20     & 91.24 $\pm$0.11   &    95.53$\pm$0.38     & 62.31$\pm$1.51           &  81.30$\pm$0.40     &   81.79$\pm$0.34    &   76.91$\pm$0.28    &   55.37$\pm$0.65    &   59.57$\pm$0.90    &  64.96$\pm$0.89   & 39.02$\pm$3.76 &57.17$\pm$0.34   &60.33$\pm$0.63   &39.44$\pm$3.29   \\
\textbf{RoBERTa-B}  &    87.64$\pm$0.12  &    87.66$\pm$0.17      & 91.94$\pm$0.07     &   97.04$\pm$0.36     & 72.45$\pm$1.02            & 82.44$\pm$0.30        &  85.13$\pm$0.15     &   81.97$\pm$0.27    &  57.39$\pm$0.63      &   63.38$\pm$0.98    &   73.84$\pm$1.61    &   52.80$\pm$3.39   &57.14$\pm$0.32   &63.92$\pm$0.67   &51.85$\pm$2.71    \\
\textbf{XLNet-B}    & 86.78$\pm$0.28  &    86.42$\pm$0.14      & 91.54$\pm$0.11 &   95.95$\pm$0.63     &  66.29$\pm$1.08         &  81.35$\pm$0.37       &    84.40$\pm$0.17   &    80.33$\pm$0.28   &    57.18$\pm$0.56    &    63.70$\pm$2.04   &    75.70$\pm$1.48   &  40.32$\pm$4.31  &56.66$\pm$0.22   &61.79$\pm$0.83   &39.93$\pm$3.91    \\
\textbf{BERT-L}  & 86.62$\pm$0.17 & 86.75$\pm$0.19 & 92.09$\pm$0.09 &   95.71$\pm$0.53 &  72.42$\pm$1.78 &     82.26$\pm$0.40 & 84.20$\pm$0.22 & 79.32$\pm$0.48 &   62.25$\pm$1.55 & 64.48$\pm$1.71 & 72.28$\pm$1.01 & 49.56$\pm$4.20  &57.19$\pm$0.29   &62.66$\pm$0.64   &49.38$\pm$3.76    \\
\textbf{RoBERTa-L} & 90.04$\pm$0.17 & 89.99$\pm$0.15 & 93.09$\pm$0.12 &  97.50$\pm$0.19 & 75.90$\pm$0.99 & 84.42$\pm$0.30 & 87.68$\pm$0.19 & 85.67$\pm$0.22 & 60.03$\pm$2.04 & 63.10$\pm$1.71 & 78.96$\pm$1.91 & 61.27$\pm$5.25 &57.77$\pm$0.29   &66.11$\pm$0.55   &58.34$\pm$4.13    \\
\textbf{XLNet-L}   &    89.48$\pm$0.20 &    89.31$\pm$0.18 &    92.90$\pm$0.14 &     97.57$\pm$0.23 &    75.75$\pm$1.22 &     83.55$\pm$0.30 & 87.33$\pm$0.18 & 84.30$\pm$0.32 & 60.46$\pm$3.25 & 67.47$\pm$2.37 & 84.26$\pm$2.14 &  62.14$\pm$3.63 &57.33$\pm$0.30   &63.56$\pm$0.70   &60.45$\pm$4.33     \\
\bottomrule
\end{tabular}%
}
\caption{Means and standard deviations of final performance on NLI datasets for all models.}

\label{tab:all_result_nli}
\end{table*}

\begin{table}[t]
\resizebox{0.47\textwidth}{!}{%
\begin{tabular}{lccc}
\toprule
\multirow{2}{*}{\bf Model}   & \multicolumn{1}{c}{Standard Dataset}                                                  & \multicolumn{2}{c}{Analysis Sets}    \\ 
\cmidrule(lr){2-2}  \cmidrule(lr){3-4} 
         &   \bf SQuAD &  \bf AddSent & \bf AddOneSent \\
\midrule
\textbf{BERT-B}     &    87.16$\pm$0.13  & 63.70$\pm$0.57 & 72.33$\pm$0.48      \\
\textbf{XLNet-B}    &    89.33$\pm$0.39  &  69.19$\pm$1.18  &   77.20$\pm$0.94    \\
\bottomrule
\end{tabular}%
}
\caption{Means and standard deviations of final F1 on SQuAD dev set for both BERT-B and XLNet-B.}
\label{tab:all_result_rc}
\end{table}

\begin{table*}[t]
\centering
\small
\begin{tabularx}{\textwidth}{rX}
\toprule
Original Context: & In February 2010, in response to controversies regarding claims in the Fourth Assessment Report, five climate scientists--all contributing or lead IPCC report authors--wrote in the journal Nature calling for changes to the IPCC. They suggested a range of new organizational options, from tightening the selection of lead authors and contributors to dumping it in favor of a small permanent body or even turning the whole climate science assessment process into a moderated ``living'' Wikipedia-IPCC. Other recommendations included that the panel employs full-time staff and remove government oversight from its processes to avoid political interference.  \\
Question: & How was it suggested that the IPCC avoid political problems?  \\
Answer: & remove government oversight from its processes \\
\midrule
Distractor Sentence 1: & It was suggested that the PANEL avoid nonpolitical problems. \\
\midrule
Distractor Sentence 2: & It was suggested that the panel could avoid nonpolitical problems by learning. \\
\bottomrule
\end{tabularx}
\caption{A highly-correlated example pair in the SQuAD-AddSent dataset based with the BERT model. This example pair have the largest covariance (0.278) among all the pairs. 
}
\label{tab:squad_example}
\end{table*}

\section{Details of Analysis Datasets}
We used the following NLI analysis datasets in our experiments: \textbf{Break NLI}~\cite{glockner2018breaking}, \textbf{SNLI-hard}~\cite{gururangan2018annotation}, \textbf{NLI Stress Test}~\cite{naik-etal-2018-stress} and \textbf{HANS}~\cite{mccoy2019right}. We use \textbf{AdvSQuAD}~\cite{jia2017advsquad} as the RC analysis dataset.

\paragraph{Break NLI.\footnote{\url{github.com/BIU-NLP/Breaking_NLI}}}
The examples in Break NLI resemble the examples in SNLI. The hypothesis is generated by swapping words in the premise so that lexical or world knowledge is required to make the correct prediction.
\paragraph{SNLI-Hard.\footnote{\url{nlp.stanford.edu/projects/snli/snli_1.0_test_hard.jsonl}}}
SNLI hard dataset is a subset of the test set of SNLI. The examples that can be predicted correctly by only looking at the annotation artifacts in the premise sentence are removed.

\paragraph{NLI Stress.\footnote{\url{abhilasharavichander.github.io/NLI_StressTest/}}}
NLI Stress datasets is a collection of datasets modified from MNLI. Each dataset targets one specific linguistic phenomenon, including word overlap (STR-O), negation (STR-NE), antonyms (STR-A), numerical reasoning (STR-NU), length mismatch (STR-L), and spelling errors (STR-S). Models with certain weaknesses will get low performance on the corresponding dataset. In our experiments, we use the mismatched set if there are both a matched version and a mismatched version. For STR-S, we follow the official evaluation script\footnote{\url{github.com/AbhilashaRavichander/NLI_StressTest/blob/master/eval.py}} to use the gram\_content\_word\_swap subset.

\paragraph{HANS.\footnote{\url{github.com/tommccoy1/hans}}}
The examples in HANS are created to reveal three heuristics used by models: the lexical overlap heuristic, the sub-sequence heuristic, and the constituent heuristic. For each heuristic, examples are generated using 5 different templates.

\paragraph{SICK.\footnote{\url{marcobaroni.org/composes/sick.html}}}
SICK is a dataset created for evaluating the compositional distributional semantic models. The sentences in this dataset come from the 8K ImageFlickr dataset and the SemEval 2012 STS MSR-Video Description dataset. The sentences are first normalized and then paired with an expanded version so that the pair can test certain lexical, syntactic, and semantic phenomena.

\paragraph{EQUATE.\footnote{\url{github.com/AbhilashaRavichander/EQUATE}}}
EQUATE is a benchmark evaluation framework for evaluating quantitative reasoning in textual entailment. It consists of five test sets. Three of them are real-world examples (RTE-Quant, NewsNLI, RedditNLI) and two of them are controlled synthetic tests (AWPNLI, Stress Test). In this work, we use EQU-NAT to denote the real-world subset and EQU-SYN to denote the synthetic tests.

\paragraph{AdvSQuAD.\footnote{Both AddSent and AddOneSent can be downloaded from \url{worksheets.codalab.org/worksheets/0xc86d3ebe69a3427d91f9aaa63f7d1e7d/}.}} AdvSQuAD is a dataset created by inserting a distracting sentence into the original paragraph. This sentence is designed to be similar to the question but containing a wrong answer in order to fool the models.

\section{Dataset Statistics}
Dataset statistics and categories for all the NLI datasets can be seen in Table \ref{tab:dataset_stats}.
Dataset statistics and categories for all the RC datasets can be seen in Table \ref{tab:dataset_stats_rc}.

\section{Training Details}
For all pre-trained transformer models, namely, BERT, RoBERTa, and XLNet, we use the same set of hyper-parameters for analysis consideration.
For NLI, we use the suggested hyper-parameters in \newcite{devlin2019bert}. The batch size is set to 32 and the peak learning rate is set to 2e-5. We save checkpoints every 500 iterations, resulting in 117 intermediate checkpoints. In our preliminary experiments, we find that tuning these hyper-parameters will not significantly influence the results. The training set for NLI is the union of SNLI~\cite{bowman2015large} and MNLI~\cite{williams2018broad}\footnote{Both SNLI and MNLI can be downloaded from \url{gluebenchmark.com}.} training set and is fixed across all the experiments. 
This will give us a good estimation of state-of-the-art performance on NLI that is fairly comparable to other analysis studies. 
For RC, we use a batch size of 12 and set the peak learning rate to 3e-5.  RC Models are trained on SQuAD1.1\footnote{\url{rajpurkar.github.io/SQuAD-explorer/}}~\cite{rajpurkar2016squad} for 2 epochs. All our experiments are run on Tesla V100 GPUs.

\section{Means and Standard Deviations of Final Results on NLI/RC datasets}
Here we provide the mean and standard deviation of the final performance over 10 different seeds on NLI and RC datasets in Table \ref{tab:all_result_nli} and Table \ref{tab:all_result_rc} respectively.

\section{High-Correlated Cases for SQuAD}
In this section, we show an example to illustrate that the high-correlated cases are similar to NLI datasets for RC datasets.
As adversarial RC datasets such as AddSent are created by appending a distractor sentence at the end of the original passage, different examples can look very similar. 
In Table~\ref{tab:squad_example}, we see two examples are created by appending two similar distractor sentences to the same context, making the predictions of these two examples highly correlated.

\end{document}